\documentclass{article}

\usepackage{arxiv}

\usepackage[utf8]{inputenc} 
\usepackage[T1]{fontenc}    
\usepackage{url}            
\usepackage{booktabs}       
\usepackage{amsfonts}       
\usepackage{nicefrac}       
\usepackage{comment}
\usepackage{graphicx}
\usepackage{microtype}      
\usepackage{lipsum}

\let\arxivabstract\abstract
\let\endarxivabstract\endabstract
\usepackage{arabtex}
\usepackage{utf8}
\let\abstract\arxivabstract
\let\endabstract\endarxivabstract

\usepackage{makecell}
\usepackage{multirow}
\usepackage{apacite}
\usepackage{natbib}
\usepackage{appendix}
\usepackage{amsmath}
\usepackage{booktabs}
\usepackage{subfig}
\usepackage{amssymb}
\usepackage{tabularx}
\usepackage{float}

\title{Automatic Classification of Arabic Literature into Historical Eras}

\author{
  Zainab Alhathloul\\
  Information and  Computer Science Department\\
  King Fahd University of Petroleum and Minerals, Dhahran 31261, Saudi Arabia \\
  \texttt{zainab-hathloul@hotmail.com} \\
 \And
  Irfan Ahmad\thanks{corresponding author} \\
  Information and  Computer Science Department\\
  King Fahd University of Petroleum and Minerals, Dhahran 31261, Saudi Arabia \\
  SDAIA--KFUPM JRC for Artificial Intelligence, Dhahran 31261, Saudi Arabia.\\
  \texttt{irfan.ahmad@kfupm.edu.sa} \\
}

\begin{document}
\maketitle
\begin{abstract}
The Arabic language has undergone notable transformations over time, including the emergence of new vocabulary, the obsolescence of others, and shifts in word usage. This evolution is evident in the distinction between the classical and modern Arabic eras. Although historians and linguists have partitioned Arabic literature into multiple eras, relatively little research has explored the automatic classification of Arabic texts by time period, particularly beyond the domain of poetry. This paper addresses this gap by employing neural networks and deep learning techniques to automatically classify Arabic texts into distinct eras and periods. The proposed models are evaluated using two datasets derived from two publicly available corpora, covering texts from the pre-Islamic to the modern era. The study examines class setups ranging from binary to 15-class classification and considers both predefined historical eras and custom periodizations.
Results range from F1-scores of 0.83 and 0.79 on the binary-era classification task using the OpenITI and APCD datasets, respectively, to 0.20 on the 15-era classification task using OpenITI and 0.18 on the 12-era classification task using APCD.

\end{abstract}


\section{Introduction}
\label{sec:intro}

Language is a dynamic system that evolves over time, influenced by semantic and lexical changes \citep{mcmahon1994understanding}. This evolution is evident in Arabic, as new words enter usage while older ones fade. Arabic literature, with a history spanning approximately 1,500 years, reflects this development and encompasses both classical and modern Arabic \citep{belinkov2019studying}.

Arabic holds official status in numerous countries, and Modern Standard Arabic is the predominant variety used in written communication. Its role as the language of the Quran and Hadith renders it central to Islamic culture. Moreover, Arabic literature extends beyond the Islamic era to include works from pre-Islamic times \citep{alrabiah2014ksucca}.

\citet{alrabiah2014ksucca} highlighted semantic change in Arabic vocabulary, demonstrating that the term \setcode{utf8}\<عادي> (pronounced “aady”) underwent shifts in meaning across different periods. In early usage, “aady” meant “old” or “an aggressor,” whereas in modern Arabic it denotes “normal.” In addition, overall word frequency is higher in Modern Arabic than in Classical Arabic \citep{alrabiah2014ksucca}. This example illustrates a semantic change across two specific periods. To demonstrate change across multiple periods, consider the word \<أدب> (pronounced “adab”), which has borne various meanings over time \citep{daif1982jahli}. Table \ref{tab:introlitmean} summarizes the evolving uses of this word in Arabic literature. In a related vein, Al-Fakhoury defined Arabic literature as “the group of written monuments in which the human mind is manifested” \citep{hanna1951history}.

\begin{table}[]
	\centering
	\caption{The semantic changes of the word \<أدب> (\textit{literature}) in the Arabic language.}
	\begin{tabular}{ll}
		\hline
		\textbf{Period}                                                                      & \textbf{Meaning}                                                                                                                                                                                                                   \\ \hline
		Pre-Islamic                                                                          & Invitation to food                                                                                                                                                                                                                 \\ \hline
		Islamic                                                                              & The moral discipline                                                                                                                                                                                                               \\ \hline
		Umayyad                                                                              & \begin{tabular}[c]{@{}l@{}}The moral discipline and they add\\ educational meaning for example:\\ teachers called preceptor.\end{tabular}                                                                                            \\ \hline
		Abbasid                                                                              &  \begin{tabular}[c]{@{}l@{}}Combined between the Islamic and\\ Umayyad meaning.\end{tabular}                                                                                                                                                                                  \\ \hline
		3th century AH                                                                       & Arab poems and their news                                                                                                                                                                                                          \\ \hline
		4th century AH                                                                       &  \begin{tabular}[c]{@{}l@{}}All the knowledge including religious\\ and non-religious knowledge.   \end{tabular}                                                                                                                                                             \\ \hline
		13th century AH                                                                      & Poems snippets and news anecdotes                                                                                                                                                                                                  \\ \hline
		\begin{tabular}[c]{@{}l@{}}from the mid of 13th \\ century AH until now\end{tabular} & \begin{tabular}[c]{@{}l@{}}Indicates two meaning: \\ 1) Any written scripts (they followed the\\ French language description of the\\ literature word)\\ 2) Pure literature that expresses a mean \\like poems and prose.\end{tabular} \\ \hline
	\end{tabular}
	\label{tab:introlitmean}
\end{table}

The foregoing examples illustrate the changes that have occurred in Arabic, particularly within Arabic literature. The origins and influence of Arabic literature have played a significant role in shaping this evolution. Although periods of human history lack written records, literature likely existed during those times \citep{hanna1951history}. For extended periods, literature was transmitted orally before being committed to writing. Linguists have noted that poetry predates artistic prose, making it among the oldest literary forms worldwide.

Automatic labeling of texts by historical period is essential for organizing literary data and tracing language change over time. The task is valuable in linguistics and history, enabling researchers to identify and analyze changes across periods \citep{niculae2014temporal}. Accurate assignment of texts to their respective periods provides insights into the language development characteristic of those times. Equally important is the ability to distinguish periods that exhibit substantial linguistic divergence from those with subtler variation, allowing analyses to focus on epochs of pronounced change while still extracting information from periods with minor shifts. Finally, such findings can aid in authenticating works attributed to specific historical eras \citep{tilahun2012dating}.

Temporal period classification is a form of text classification. Text classification has been introduced for various purposes, such as sentiment analysis and topical categorization \citep{elnagar2020arabic,farha2019mazajak,alhawarat2015processing}. However, temporal period classification has received less attention than other text-classification tasks, particularly in Arabic. Only two studies in the literature have focused on classifying poems into different eras. Accordingly, this study specifically addresses the temporal period classification of Arabic literature, encompassing both poetry and non-poetry texts.

\subsection{Research Objectives}

The primary objective of this study is to assess the extent to which Arabic literature can be automatically classified into historical eras. The objective comprises the following sub-objectives:
\begin{enumerate}
\item Investigate the automatic classification of Arabic literature into linguist-defined eras using neural-network and deep-learning models.
\item Investigate the automatic classification of Arabic literature into custom time periods using neural-network and deep-learning models, and compare the results with classification into linguist-defined eras.
\item Investigate the performance of classifying Arabic \emph{poetry} into different eras and compare with the classification of \emph{non-poetry} texts.
\item Investigate the impact of authorial style on automatic era-classification tasks.
\end{enumerate}

Guided by these objectives, a series of experiments was conducted on two publicly available datasets, yielding several insights and conclusions.

\subsection{Primary Contributions}

To the best of current knowledge, the categorization of Arabic literary texts into distinct eras has not been systematically explored, underscoring the need to investigate automatic temporal classification. This paper addresses this gap and makes the following contributions:

\begin{enumerate}
\item Proposes an Arabic-era classification method that utilizes two datasets derived from two publicly available text corpora.
\item Examines classification under both predefined historical eras and custom-defined temporal periods.
\item Conducts two experimental protocols to assess the impact of authorial style: in one, authors in the evaluation set also appear in the training set; in the other, the splits are author-disjoint, with evaluation-set authors absent from the training set.
\item Evaluates both fully connected feedforward artificial neural networks (ANNs) and recurrent neural networks (RNNs) using two different text-tokenization approaches.
\end{enumerate}

\subsection{Structure of the article}

The rest of the paper is organized as follows: Section \ref{sec:background} presents the background of the study, including an introduction to the concept of literature eras in Arabic and the underlying motivation for this research. Section \ref{sec:related} presents a comprehensive literature review on the topic. Section \ref{sec:datasets} describes the datasets utilized in the experiments. Section \ref{sec:method} outlines the methodology employed in this study. Section \ref{sec:exp} presents and discusses the experimental findings. Finally, Section \ref{sec:conc} presents the conclusions drawn from the study.

\section{Background}
\label{sec:background}

In Arabic literary studies, numerous eras have been proposed, often with debated and overlapping boundaries. Nevertheless, certain eras—such as the pre-Islamic and Islamic—are consistently recognized.

\citet{daif1982jahli} categorized Arabic literature into five eras, reflecting the views of most historians: pre-Islamic, Islamic, Abbasid, Aldoul wa al-emarat, and modern. Similarly, \citet{hanna1951history} proposed a comparable five-era division—pre-Islamic, Islamic, Abbasid, Turkish, and modern—structured around three historical revivals: the pre-Islamic and Umayyad, Abbasid, and modern renaissances.

Previous research commonly identifies the first three eras as pre-Islamic, Islamic, and Abbasid. Within the Islamic era, many studies further divide it into the Islamic and Umayyad periods. However, the literature is inconsistent regarding the end date of the Abbasid era. Notably, the schemes of \citet{daif1982jahli} and \citet{hanna1951history} conflict and display overlapping boundaries. In response, \citet{ghoniem2020literature} expanded the framework to six eras: pre-Islamic, Islamic, Abbasid, Aldoul wa al-emarat, Ottoman, and modern. Table \ref{tab:the6ErasPeriods} presents the start and end dates of each era according to the Hijri calendar, with Aldoul wa al-emarat being the longest and the modern era the shortest.

\begin{table}
	\centering
	\caption{The periods of six literature eras by Ghoniem \cite{ghoniem2020literature}.}
	\begin{tabular}{@{}lr@{}}
		\toprule
		\textbf{Era}    & \textbf{Period}     \\ \midrule
		Pre-Islamic         & 150 years before 1 AH \\ \hline
		Islamic             &  1 AH – 132 AH      \\ \hline
		Abbasid            & 132 AH – 334 AH    \\ \hline
		\textit{Aldoul wa al-emarat} & 334 AH –   923 AH  \\ \hline
		Ottoman           & 923 AH – 1335 AH    \\ \hline
		Modern            & 1335 AH –  till date \\ \bottomrule
	\end{tabular}
	\label{tab:the6ErasPeriods}
\end{table}

The division of Arabic literary eras remains a complex, open area of inquiry for historians and literary critics. Accordingly, this study adopts the recent scheme of \citet{ghoniem2020literature}, which minimizes overlaps and provides a more comprehensive categorization.

\section{Related Work}
\label{sec:related}

This section reviews studies that date text documents by their exact date or by the period to which they belong. Most research employs machine-learning approaches, although some studies adopt statistical methods; consequently, the problem is modeled in different ways across the literature.

Much of the prior work focuses on English texts, with only three studies specifically addressing Arabic. Related efforts also exist for other languages, such as Dutch \citep{jong2005temporal}. Accordingly, the review is organized into two categories: studies on Arabic and studies on English, with work in other languages noted for context.

\subsection{Works Related to Arabic Text}
\label{arabicmethods}

\citet{abbas2019classification} classified Arabic poems into four eras—pre-Islamic, Umayyad, Abbasid, and Andalusian—using a dataset of approximately 58{,}000 poems spanning the 6th to the 21st centuries. The experiments, conducted with WEKA, evaluated eight classifiers, including k-nearest neighbors (k-NN), multinomial Naive Bayes, a bagging classifier, and five others. Multinomial Naive Bayes outperformed the alternatives, achieving an accuracy of 70.21\% and an F1-score of 68.8\%.

\citet{belinkov2019studying} introduced a periodization algorithm, word-embedding-based neighbor clustering (WENC), designed to automatically identify periods of language development to support Arabic learning. The study utilized the Open Islamicate Texts Initiative (OpenITI) corpus and a preprocessed version of the data. The authors enhanced the text-reuse detection algorithm proposed by \citet{porat2018identification} for OpenITI, where text reuse includes boilerplate passages, extremely frequent short phrases, and approximate matches. WENC relies on word embeddings to cluster periods with similar language use. The study identified several distinct periods in the development of Arabic and noted that lexical items persist longer in Arabic literature than in English. Three principal eras—early, middle, and late—were reported \citep{belinkov2019studying}.

\citet{orabi2020classical} classified Arabic poetry into the same eras as \citet{abbas2019classification} and added a modern era. Their dataset augmented \citet{abbas2019classification} by 2{,}000 poems. The study proposed a convolutional neural network (CNN) and examined several experimental class setups. Each setup included a baseline and a deep-learning model. The baseline used FastText word embeddings trained on an unlabeled Arabic poetry corpus, whereas the deep-learning model employed a CNN. An F1-score of 0.914 was reported for the binary classification task.

\citet{RUMA2022100111} presented a deep-learning system for classifying poems by the Persian poet Hafez into different eras (four to six) within his lifetime. Distributed Bag of Words (DBOW) and Distributed Memory (DM; similar to the Continuous Bag of Words model, CBOW) representations were extracted from the poems and fed into various RNN architectures. The LSTM model achieved the best F1-score.

\citet{jimaging8030060} addressed the dating of historical handwritten manuscript images. ResNet was used for feature extraction, alongside additional image features such as Gabor filters and histogram of oriented gradients (HOG). The features were fused in a hierarchical manner.

\subsection{Works Related to English Texts}

\citet{colgrove2010literary} investigated genre classification across four literary movements using the Gutenberg corpus, focusing on English fiction. Various features were extracted to train a maximum-entropy classifier. The authors evaluated multiple experimental setups, including a standard “general” setup, which achieved an F1-score of 0.80 \citep{colgrove2010literary}.

\citet{kumar2011supervised} proposed a supervised language model for dating texts. The model represents documents with histograms and compares their associated language models using Kullback–Leibler divergence. The Project Gutenberg corpus and Wikipedia biographies were used for training and evaluation, alongside a baseline model for comparison. The proposed approach outperformed the baseline, reducing the median error from 50 to 23 \citep{kumar2011supervised}.

\citet{mihalcea2012word} proposed classifying word usage by historical period, using the Google Books N-Gram corpus. Experiments were conducted on word-usage data sampled at three time points (1800, 1900, and 2000 CE), integrating a set of features into a Naive Bayes classifier. The model achieved an 18.5\% improvement over a 43\% baseline.

\citet{vstajner2013stylistic} examined stylistic change in texts from the 17th to the 20th century. A quantitative analysis was used to automatically extract stylistic features from the Colonia Corpus, and multiple experiments classified texts into five class sets. WEKA implementations of Naive Bayes (NB), support vector machines (SMO), JRip, and J48 were evaluated; NB and SMO obtained the highest F-measures of 0.92 \citep{vstajner2013stylistic}.

\citet{popescu2015semeval} introduced the Diachronic Text Evaluation (DTE) task at SemEval 2015, comprising three subtasks (T1–T3) based on a newspaper dataset with fine, medium, and coarse temporal intervals. \citet{salaberri2015ixagroupehudiac} addressed all three tasks using temporal text classification for T1 and T2 and a phrase-relevance method for T3. Four approaches—year-entity detection, Wikipedia entity linking, Google N-grams, and language-change features—were explored for predicting a text’s period. The final decision strategy achieved the highest overall precision, and among individual methods, year-entity detection yielded the best precision. The best DTE score (0.62) was obtained on T2 \citep{salaberri2015ixagroupehudiac}.

\citet{zampieri2015ambra} proposed the Anachronism Modeling by Ranking (AMBRA) system for T1 and T2 of DTE, employing the pairwise ranking approach of \citet{niculae2014temporal} to assign news documents to temporal intervals and leveraging a diverse feature set. AMBRA outperformed competing systems with a DTE score of 0.87 \citep{zampieri2015ambra}.

\citet{szymanski2015ucd} participated in T2, expanding the three competition intervals by adding a fourth period. An SVM classifier was trained with four stylistic features, supplemented by features extracted from GSN. The best accuracy—73.3\%—was achieved for the 50-year setup \citep{szymanski2015ucd}.

\section{Datasets}
\label{sec:datasets}

Fewer resources exist for historical Arabic text than for other languages, especially English \citep{alrabiah2014ksucca}. Moreover, most Arabic corpora comprise Modern Arabic rather than historical literature \citep{belinkov2019studying}. Nevertheless, several corpora provide historical Arabic literature relevant to this study.

Two corpora are used in the experiments: the OpenITI corpus \citep{maxim2019romanov} and the Arabic Poem Comprehensive Dataset (APCD) \citep{PCD2018}. OpenITI is a large-scale collection of Arabic texts from diverse authors and books, whereas APCD consists of Arabic poetic verses spanning several eras.

For both corpora, the data were split into training, validation, and test sets. The training–validation portion comprised 85\% of the data, with the remaining 15\% reserved for testing. Within the training portion, 15\% was held out for validation.

Two dataset variants were prepared for each corpus. In the first (author-disjoint) variant, authors were partitioned across training, validation, and test such that no samples from evaluation-set authors appeared in the training set. In the second (merged-author) variant, authors could appear in both training and evaluation splits. These complementary protocols provide insight into the effect of authorial style on model performance. Under the author-disjoint setting, the aim is to encourage the model to learn period-level stylometric signals rather than author-specific style.

\subsection{OpenITI Dataset For Temporal Text Classification}
\label{openITIDataset}

\citet{belinkov2019studying} introduced the OpenITI corpus, which is publicly available \citep{maxim2019romanov}. OpenITI is a large historical corpus of Arabic texts (and other languages) with approximately 1.5 billion words. Collected across different periods, it provides a freely available resource for religious and literary texts. Al-Maktaba al-Shamela (Shamela) was the primary source \citep{belinkov2019studying}. Owing to the corpus’s size and the exceptional length of many documents, subsets were created by extracting samples of up to 100 words. Because each author may have published multiple books, the number of samples was standardized across an author’s books to ensure broader coverage of the original corpus. When multiple versions of a book were available, the first version was used.

The OpenITI corpus comprises numerous books with an average length of 176{,}718 words. To avoid potentially irrelevant content (e.g., prefaces, tables of contents), the first few sentences were omitted before sampling. Two non-Arabic books were also excluded.

Different class setups were examined, yielding two experimental sets. The first set classified texts into well-defined eras: a binary scheme, the three eras proposed by \citet{belinkov2019studying}, and the five eras proposed by \citet{ghoniem2020literature}. The binary scheme collapses the five eras into modern versus classical, with the classical category comprising the four non-modern eras. Table \ref{tab:openITI5ErasPeriods} lists the periods for the five-era setup. Note that the pre-Islamic era was excluded from this five-era setup due to limited available texts; pre-Islamic texts were discarded only in this setup.

\begin{table}
	\centering
	\caption{The periods for the five literature eras setup in OpenITI dataset.}
	\begin{tabular}{@{}lr@{}}
		\toprule
		\textbf{Era}    & \textbf{Period}   \\ \midrule
		Islamic             & 50 – 175 AH      \\ \hline
		Abbasid            & 175 – 400 AH     \\ \hline
		\textit{Aldoul wa al-emarat} & 400 –   950 AH   \\ \hline
		Ottoman           & 950 – 1350 AH    \\  \hline
		Modern            & 1350 –   1450 AH \\ \bottomrule
	\end{tabular}

	\label{tab:openITI5ErasPeriods}
\end{table}

Table \ref{tab:threeperiodsbyben} summarizes the three eras and their corresponding ranges proposed by \citet{belinkov2019studying}, listing the original start and end years. Because the final year of the early era is reported as either 200 or 300 AH, the boundary was standardized to 300 AH. Likewise, the middle era, originally ending around 1300 AH, was set to end at 1300 AH.

In addition, custom periodizations of 300-, 200-, and 100-year bins were examined, yielding five, eight, and fifteen classes, respectively. These bins were used as exploratory alternatives to historically grounded eras, allowing assessment of the sensitivity of performance to periodization design and visualization of confusion patterns at coarser versus finer temporal resolutions.

\begin{table}[]
	\centering
	\caption{The three periods for the Arabic language development proposed by \cite{belinkov2019studying} in the OpenITI dataset.}
	\begin{tabular}{@{}ll@{}}
		\toprule
		\textbf{Era} & \textbf{Period}                           \\ \midrule
		Early period  & from 1 AH to 200/300 AH           \\ \hline
		Middle period & from 200/300 AH to around 1300 AH \\ \hline
		Late period   & from 1300 AH to modern days       \\ \bottomrule
	\end{tabular}
	\label{tab:threeperiodsbyben}
\end{table}

Table \ref{tab:openITIdatasetSizes} reports the sample distribution for all OpenITI dataset setups, along with the total size for each. The binary era setup contains the largest number of samples, whereas the 200-year custom periodization yields the fewest.

\begin{table*}[t]
	\centering
	\caption{The number of samples for all the class setups in OpenITI dataset.}
	\begin{tabular}{@{}lrrrr@{}}
		\toprule
		\textbf{Setup} & \multicolumn{1}{c}{\textbf{Training}} & \multicolumn{1}{c}{\textbf{Validation}} & \multicolumn{1}{c}{\textbf{Test}} & \multicolumn{1}{c}{\textbf{Totals}} \\ \midrule
		Binary literature eras        & 54,400                                   & 9,602                                     & 11,296                                & 75,298   
		\\ \hline
		3-Eras as Belinkov et al. \cite{belinkov2019studying} proposed      & 24,100                                   & 4,242                                      & 4,988                               & 33,330   
		\\ \hline
		Five literature eras            & 34,000                                   & 6,000                                      & 7,055                               & 47,055                                 \\ \hline
		300-years custom period (5-class)        & 34,000                                   & 6,000                                      & 7,055                               & 47,055    
		\\ \hline
		200-years custom period (8-class)       & 24,000                                   & 4,232                                      & 4,982                                & 33,214
		\\ \hline
		100-years custom period (15-class)       & 30,000                                   & 5,292                                      & 6,226                                & 41,518   
		\\
		\bottomrule          
	\end{tabular}
	\label{tab:openITIdatasetSizes}
\end{table*}

\subsection{APCD Dataset For Temporal Poem Classification}

\citet{PCD2018} compiled the APCD corpus from two primary resources: the Poetry Encyclopedia \footnote{https://poetry.dctabudhabi.ae/} and Diwan \footnote{https://www.aldiwan.net/}. Each record includes the poem verse, poet name, metre, and era. The dataset spans 3{,}701 poets across 12 eras—pre-Islamic, Mukhadramayn, Islamic, Umayyad, Mamluk, Abbasid, Ayyubid, Ottoman, Andalusian, the period between Umayyad and Abbasid, Fatimid, and modern—and contains approximately 1.8M verses.

ll verses by a given author were concatenated into a single file. Samples were then extracted to create a balanced dataset while maintaining author separation. Because individual verses lack poem titles, an author’s verses may originate from different poems.

Given the corpus size, subsets were generated to examine the effect of sample size on classification performance. Configurations were created with 1 to 16 verses per sample.

Several class setups were investigated: binary, five-era, and twelve-era. The binary setup contrasts classical versus modern eras, following the approach used for OpenITI. The five-era setup adopts the division of \citet{ghoniem2020literature}—pre-Islamic, Islamic, Abbasid, Ottoman, and modern—aligning with the five OpenITI literary eras described in Section \ref{openITIDataset}. Unlike the OpenITI dataset, the \textit{Aldoul wa al-emarat} era was excluded and the pre-Islamic era included. The twelve-era setup retained all eras from the original APCD corpus.

In the five-era setup, the Islamic era combines the Islamic and Umayyad periods under the adopted literary-era division, and therefore differs from the twelve-era setup, which treats them separately.

For each setup, verses-per-sample values from 1 to 16 were evaluated. The maximum was capped at 16 owing to data scarcity in some eras. In the twelve-era setting only, the Islamic era was excluded for configurations with more than four verses per sample because of limited data, yielding an eleven-era variant.

Table \ref{tab:poemDataSizes} reports the sample distributions for each APCD dataset. The second column lists the verses-per-sample range for each setup. For a given setup, dataset size was adjusted to achieve class balance; as the number of verses per sample increased, the overall dataset size decreased.

\begin{table*}[t]
	\centering
		\caption{The samples distribution of all the class setups in APCD dataset.}
	\begin{tabular}{@{}lccccc@{}}
		\toprule
		\textbf{Setup}                                                                    & \textbf{\begin{tabular}[c]{@{}c@{}}Number of \\ Verse per Sample\end{tabular}} & \textbf{Training} & \textbf{Validation} & \textbf{Test} & \textbf{Total} \\ \midrule
		\multirow{2}{*}{\begin{tabular}[l]{@{}c@{}}Binary literature eras\end{tabular}} & 1 -- 8                & 24,000              & 4,242               & 4,988         & 33,330         \\ \cline{2-6}
		& 12 -- 16                                                                        & 16,000              & 2,819               & 3,318         & 22,137        \\ \hline
		\multirow{3}{*}{\begin{tabular}[l]{@{}c@{}}Five literature eras\end{tabular}}      & 1 -- 8                                                                          & 15,000              & 2,645                & 3,110          & 20,755          \\ \cline{2-6}
		& 12 -- 16                                                                        & 8,498             & 1,498               & 1,761         & 11,757          \\ \hline
		\multirow{2}{*}{11-eras}                                                                                    & 5 -- 10                                                                         & 18,700            & 3,300               & 3,882          & 25,882         \\ \cline{2-6}
	     & 11 -- 16                                                                         & 13,200            & 2,329               & 2,740          & 18,269         \\
		\hline
		\multirow{2}{*}{12-eras}                                                                                    & 1 -- 2                                                                          & 20,400            & 3,600              & 4,235         & 28,235        \\ \cline{2-6}
		 & 3 -- 4                                                                          & 9,600           & 1,694              & 1,993         & 13,287         \\ 
		 \bottomrule
	\end{tabular}
	\label{tab:poemDataSizes}
\end{table*}

\section{Methodology}
\label{sec:method}

This section describes the methodology for classifying Arabic texts by temporal period. The task is formulated as text classification. To assign texts to periods, the approach centers on neural networks, including deep-learning models. Neural networks have shown promising results for Arabic natural language processing (ANLP) and offer automated feature learning, reducing manual feature engineering—a critical step in machine learning \cite{chollet2017deep}. This makes them well suited to temporal classification in a morphologically rich language such as Arabic.

The experiments focus on two neural architectures: fully connected artificial neural networks (ANNs) and recurrent neural networks (RNNs), each evaluated under multiple configurations. In addition, convolutional neural networks (CNNs) and logistic regression are included as secondary baselines.

\subsection{Data Pre-Processing}
\label{ssec:preprocess}

Preparing the data was essential to ensure proper model learning. Raw text typically contains extraneous information unrelated to the task, which can lead to overfitting \cite{uysal2014impact}. Accordingly, some content was removed and other elements reduced (e.g., via lemmatization). The preprocessing steps used are defined below:

\begin{enumerate}

    \item \textbf{Data cleaning:} The datasets contained non-Arabic digits, hypertext markup language (HTML) tags, symbols, and punctuation. Regular expressions (RegEx) were used to remove content unrelated to the task, including symbols, punctuation, non-Arabic digits, and HTML tags. For example, symbols such as '|' and '\#' were removed.
 
	\item \textbf{Normalization:} The different forms were treated as separate words, which increased the vocabulary size. Overfitting can be avoided by reducing and converting them into a more unified sequence as follows:
	\begin{itemize}
		\item Diacritics like "\<ِ ٍ ُ>"  were removed. 
		\item \textit{Kashida} "\<ـ>" characters were removed. \textit{Kashida} is used for aesthetic reasons. It is used to make the words look longer without affecting the meaning. For example, using Kashida in the word "\<الرسول>" can render the word as "\<الرســـــول>".
    \end{itemize}

	\item \textbf{Remove stop words:} Every language includes a set of commonly used function words (e.g., prepositions). Such stop words appear across most samples and eras, contribute little discriminative power, and may increase model complexity and overfitting. Accordingly, Arabic stop words were removed using the Natural Language Toolkit (NLTK) list\footnote{https://www.nltk.org/}.
	
	\item \textbf{Lemmatization:} Lemmatization performs morphological analysis to map inflected forms to their base form (lemma), thereby reducing vocabulary size and mitigating overfitting. Farasa, a widely used tool for Arabic text processing proposed by Darwish et al. \cite{darwish2016farasa} and adopted by Belinkov et al. \cite{belinkov2019studying}, was used to lemmatize the data\footnote{https://alt.qcri.org/farasa/}. Because lemmatization is computationally intensive, it was applied where it was most informative—namely, to the five-era classification for both datasets.
    \end{enumerate}
 
	Table \ref{tab:preprocessexamples} illustrates an example before and after text preprocessing. The first row contains the original text. The second row shows the result after removing diacritics. In the original sample, most words bear diacritics—some have a single mark, whereas others have more than one. The third row presents the sample after stop-word removal; the words \<فيه> and \<في> were removed. Finally, the fourth row shows the sample after lemmatization. The word \<السؤال> (the question) and it converted to its lemma \<سؤال> (question). Lemmatization was not applied in all experiments; instead, experiments were conducted with and without lemmatization as an ablation study to assess its effect.

\begin{table}[]
	\centering
	\caption{Example of pre-processed sample using remove stop words and lemmatization.}
	\begin{tabular}{@{}lr@{}}
		\toprule
		\textbf{The Applied Pre-processing}                  & \textbf{Example} \\ \midrule
		The original sample      &  \makecell{   \<تلوذُ به  الأكابرُ في صغارٍ > \\ \<وترجو فيه مَقبولَ السؤالِ>}  \\ \hline
		Normalization     & \makecell{     \<تلوذ به الأكابر في صغار  > \\ \<وترجو فيه مقبول السؤال>}   \\ \hline
		Remove stop words &    \makecell{   \<تلوذ الأكابر صغار > \\ \<وترجو مقبول السؤال>}\\ \hline
		Lemmatization     &  \makecell{    \<تلوذ ب أكبر في صغير > \\ \<رجا في مقبول سؤال>} \\ \bottomrule
	\end{tabular}
	\label{tab:preprocessexamples}
\end{table}

\subsection{Era Classification Using Fully-Connected Feed-Forward Networks}
A network of fully connected layers was investigated as a baseline model for temporal classification. Input data were prepared and preprocessed as described in Section \ref{ssec:preprocess}. For tokenization, the following feature representations were used:

\begin{enumerate}
\item \textbf{Bag-of-Words (BoW):} converts texts into sparse vectors whose dimensionality equals the dataset vocabulary size, where each index indicates the presence or absence of a specific word.
\item \textbf{Term frequency–inverse document frequency (TF–IDF):} a sparse representation similar to BoW, but weighting words by their importance based on their counts across documents (samples in this case).
\end{enumerate}

\begin{figure}
	\centering
	\includegraphics[width=3.2in]{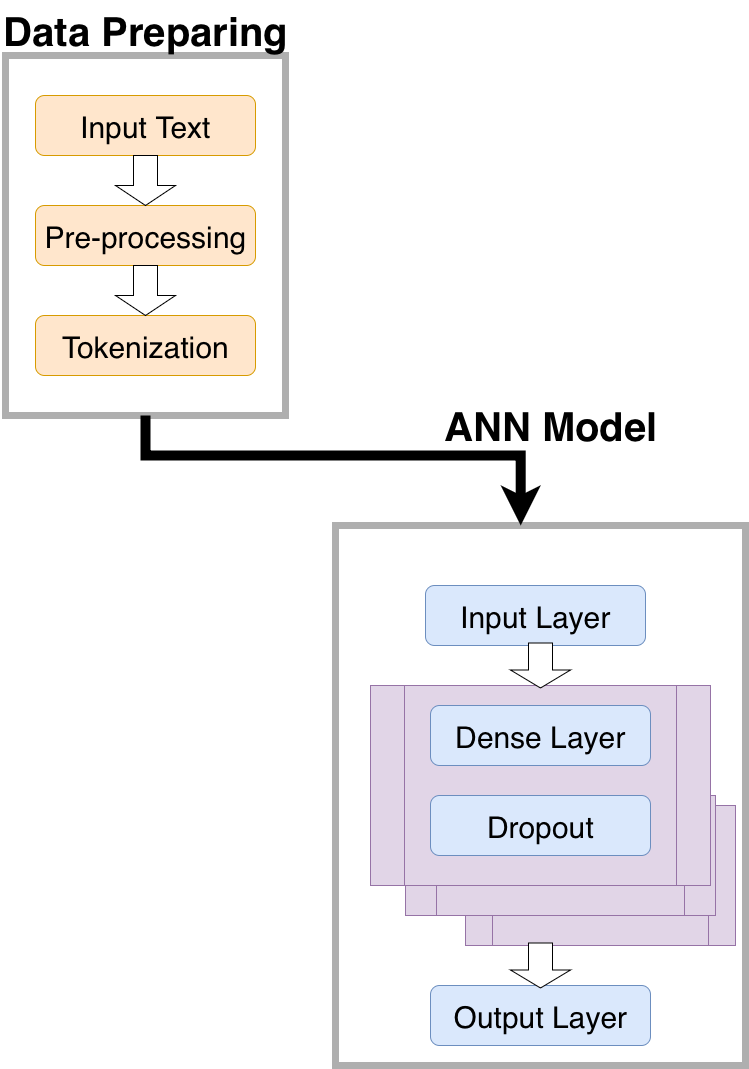}
	\caption{Illustration for the architecture of the ANN model for text-era classification.}
	\label{fig:annmethod}
\end{figure}

Figure \ref{fig:annmethod} illustrates the methodology of the ANN model for classifying Arabic texts into time periods.

The ANN model architecture comprises several layers, described below:
\begin{itemize}
\item \textbf{Input layer:} Receives the input vector representing a sample and transfers it to the subsequent layer.
\item \textbf{Dense layer:} Each neuron is connected to all neurons in the next layer. This layer applies a nonlinear transformation of the features; ReLU is used as the activation function.
\item \textbf{Dropout layer:} Used to mitigate overfitting. Dropout is a form of regularization that randomly drops units in a layer by setting their activations to zero.
\item \textbf{Output layer:} Produces the final predictions. The activation function depends on the number of classes: \textit{sigmoid} is used for binary classification, and \textit{softmax} is used for multiclass classification.
\end{itemize}
The ANN model was tested with different numbers of layers. In each case, a fully connected (dense) layer is followed by a dropout layer, as indicated by the purple box in Figure \ref{fig:annmethod}.

\subsection{Era Classification Using Recurrent Neural Networks}

\begin{figure}
\centering
\includegraphics[width=3.2in]{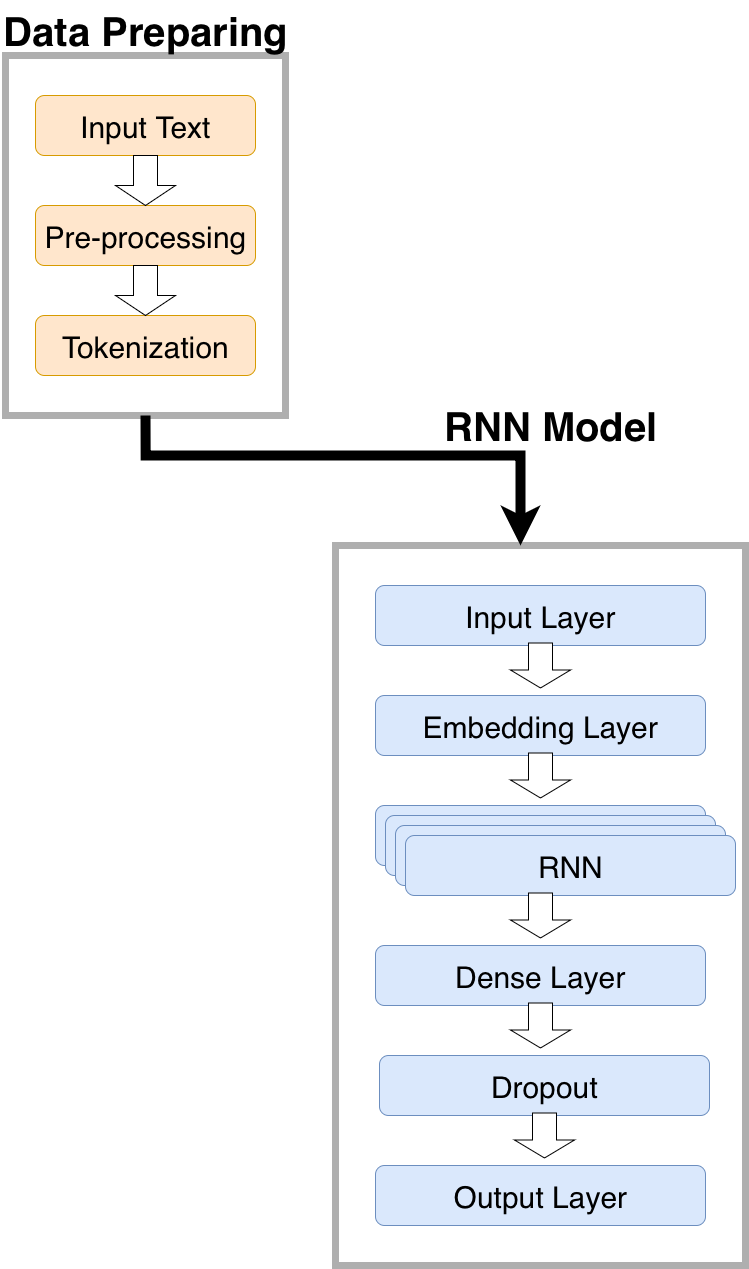}
\caption{RNN methodology structure.}
\label{fig:rnnmethod}
\end{figure}

An RNN model was investigated for classifying Arabic text into temporal periods. Figure \ref{fig:rnnmethod} depicts the RNN methodology used for the task. The input data were prepared before model ingestion, with preprocessing as described in Section \ref{ssec:preprocess}. During tokenization, two approaches were explored—word-level and character-level—resulting in the following schemes:

\begin{enumerate}
\item \textbf{Words as tokens:} A sequence of words was used as input. Each word was represented by a unique integer, yielding a dictionary of all unique words in the dataset after preprocessing. The dictionary size varied with the dataset vocabulary.
\item \textbf{Characters as tokens:} Character sequences were used as input tokens. The same procedure as the word-token scheme was followed, except that characters were represented instead of words.
\end{enumerate}

Because sample lengths differ under both schemes, the maximum sequence length (in tokens) was computed, and post-padding was applied to unify sample lengths.

The RNN architecture in Figure \ref{fig:rnnmethod} comprises several layers. The input, dropout, dense, and output layers are as described for the ANN methodology. The embedding layer was initialized with random weights and learned word/character embeddings over the training set during optimization. The recurrent layer was instantiated as one of three types: long short-term memory (LSTM), gated recurrent unit (GRU), or bidirectional GRU (BiGRU). ReLU was used as the activation function in the preceding nonrecurrent layers.

Two architectural setups were evaluated: word-level and character-level models. Each has distinct advantages and limitations. The word-level approach benefits from shorter input sequences but suffers from the out-of-vocabulary (OOV) problem, wherein unseen words cannot be directly handled \cite{lee2017fully}. The character-level approach avoids OOV issues and uses a smaller input alphabet, offering greater flexibility in composing unseen words; however, it yields longer input sequences that typically require deeper or more expressive architectures to capture sufficient context.

\subsection{Evaluation Metrics}

Accuracy, precision, recall, and F1-score were used to evaluate the temporal text classification models. Macro-averaging was applied to give equal weight to each class. Confusion matrices were visualized as heat maps to summarize misclassifications. In addition, the performance of two classifiers trained on the same dataset was compared. Statistical significance was assessed using the difference-between-two-proportions test of Dietterich et al. \cite{dietterich1998approximate}, with a 95\% confidence level.

\section{Experiments, Results, and Discussions}
\label{sec:exp}

This section presents the experiments conducted to classify Arabic texts by temporal period. ANNs and RNNs were used for the task, trained on the OpenITI and APCD datasets. For each dataset, several class setups were examined, consisting of either predefined historical periods or custom periodizations. The five-literary-eras setup serves as the predefined example, whereas the 300-year custom periodization represents the custom case. To select model configurations for each dataset, model design was first tuned on the five-era setup; the resulting best design was then applied to the remaining setups.

Removing stop words and lemmatization are common in ANLP and can improve performance; their effectiveness was therefore evaluated in the experiments. 

Experiments were implemented in Google Colab. Keras with a TensorFlow backend was used to train the models. In addition, {\itshape scikit-learn} libraries were employed for tokenization techniques and traditional machine-learning baselines. Binary cross-entropy was used for binary-class models, and sparse categorical cross-entropy for multiclass models. All models were trained for 10 epochs. Key hyperparameters for the ANN and RNN models are summarized in Tables \ref{tab:hyperparams-ANN} and \ref{tab:hyperparams-RNN}, respectively.

\begin{table}[htbp]
\caption{Important hyperparameters for the ANN model.}
\label{tab:hyperparams-ANN}
\centering
\begin{tabularx}{\linewidth}{@{}l X@{}}
\toprule
\textbf{Hyperparameter} & \textbf{Value} \\
\midrule
Dense Layer Neurons & 32 \\
Activation & ReLU \\
Dropout Rate & 0.7 \\
Optimizer & RmsProp (Adam for the APCD dataset) \\
Learning regime & 10 epochs with early stopping and saving the best model based on the validation-set accuracy. \\
Batch Size & 512 samples \\
Vocabulary Size (V) & 15000 \\
\bottomrule
\end{tabularx}
\end{table}

\begin{table}[htbp]
\caption{Important hyperparameters for the RNN model.}
\label{tab:hyperparams-RNN}
\centering
\begin{tabularx}{\linewidth}{@{}l X@{}}
\toprule
\textbf{Hyperparameter} & \textbf{Value} \\
\midrule
RNN type & BiGRU with 32 neurons, two layers deep \\
Dropout Rate & 0.7 \\
Optimizer & RmsProp \\
Learning regime & 10 epochs with early stopping and saving the best model based on the validation-set accuracy. \\
Batch Size & 128 samples \\
Vocabulary Size (V) & 20{,}000 (80{,}000 for the APCD dataset) \\
\bottomrule
\end{tabularx}
\end{table}

Results across datasets are reported on the test set in terms of accuracy, precision, recall, and F1-score. Performance is presented both with and without author merging. Retaining stop words yielded better performance than removing them; accordingly, all reported results exclude stop-word removal. Finally, lemmatization improved some results, which are reported where applicable.

\subsection{Temporal Classification with Authors Separate}
\subsubsection{OpenITI Dataset Results}

The ANN model is evaluated first under author-disjoint splits (separate authors in the training, validation, and test sets). On OpenITI, the ANN outperforms the other models across all class setups (Table \ref{tab:bestopenitiannresults}). Binary and three-era configurations yield comparable test performance, whereas accuracy declines once the number of classes exceeds five, reflecting the difficulty of imposing sharp boundaries on a gradual literary continuum; nevertheless, performance remains above chance. Because the datasets are class-balanced, the various metrics are broadly similar. Per-class results for the five-era setup are summarized in Table \ref{tab:perclass-metrics-openITI}: overall performance is modest (macro precision/recall/F1 $\approx$ 44.7/43.8/43.6\%), with the Abbasid era strongest (F1 $\approx$ 50.6\%, highest recall 53.5\%). The edge eras (Islamic and Modern) are the most difficult to detect (lowest recall: 33.7\% and 37.3\%), though the Islamic era attains the best precision (51.7\%), indicating conservative but relatively accurate predictions.

\begin{table*}[]
	\centering
	\caption{The best achieved results in OpenITI dataset using the ANN model with authors separated. The accuracy values are shown along with statistical significance interval at 95\% confidence level.}
	\begin{tabular}{@{}lcccc@{}}
		\toprule
		\textbf{Class Setup}                                                         & \multicolumn{1}{l}{\textbf{Accuracy}} & \multicolumn{1}{l}{\textbf{Precision}} & \multicolumn{1}{l}{\textbf{Recall}} & \multicolumn{1}{l}{\textbf{F1-Measure}} \\ \midrule
		\begin{tabular}[c]{@{}l@{}}Binary literature eras\end{tabular}        & 0.674 $\pm{0.0087}$                                & 0.690                                  & 0.675                               & 0.668                                  \\ \hline
		3-eras                                                                  & 0.664 $\pm{0.0132}$                                   & 0.661                                  & 0.664                               & 0.662                                  \\ \hline
		\begin{tabular}[c]{@{}l@{}} Five literature eras \end{tabular}            & 0.436 $\pm{0.0116}$                                   & 0.447                                  & 0.438                               & 0.436                                  \\ \hline
		
		\begin{tabular}[c]{@{}l@{}}300-years custom period\\ (5-class)\end{tabular}  & 0.455  $\pm{0.0116}$                                 & 0.455                                  & 0.459                               & 0.456                                  \\ \hline
		\begin{tabular}[c]{@{}l@{}}200-years custom period\\ (8-class) \end{tabular} & 0.329  $\pm{0.0129}$                                 & 0.328                                  & 0.334                               & 0.329                                  \\ \hline
		\begin{tabular}[c]{@{}l@{}}100-years custom period\\ (15-class)\end{tabular} & 0.203  $\pm{0.0098}$                                 & 0.202                                  & 0.208                               & 0.202                                  \\ \bottomrule
	\end{tabular}
	\label{tab:bestopenitiannresults}
\end{table*}

\begin{table}[htbp]
\caption{Summary of the per-class performance for the 5-era classification using the OpenITI dataset.}
\label{tab:perclass-metrics-openITI}
\centering
\begin{tabular}{lcccc}
\hline
\textbf{Class} & \textbf{Accuracy (\%)} & \textbf{Precision (\%)} & \textbf{Recall (\%)} & \textbf{F1-score (\%)} \\
\hline
Islamic & 33.70 & 51.67 & 33.70 & 40.79  \\
Abbasid & 53.50 & 47.97 & 53.50 & 50.59  \\
Aldoul wa al-emarat & 46.24 & 36.52 & 46.24 & 40.81  \\
Ottoman & 48.12 & 42.38 & 48.12 & 45.07  \\
Modern & 37.26 & 44.82 & 37.26 & 40.69  \\
\hline
\end{tabular}
\end{table}

Table \ref{tab:openitirnnresults} reports results for the word-level RNN model. Binary classification is easier than multiclass classification; accordingly, accuracy decreases as the number of classes increases. For larger class counts, the ANN outperforms the RNN.

\begin{figure}
\begin{center}
\includegraphics[width=0.45\textwidth]{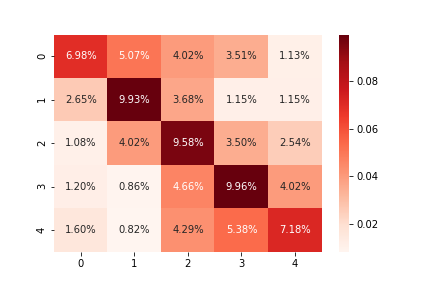}
\caption{The confusion matrix for the five literature eras classification using the OpenITI dataset and the ANN model.}
	\label{fig:cvopenitiann5}
\end{center}
\end{figure}

Figure \ref{fig:cvopenitiann5} shows the five-era confusion matrix for the ANN. More than half of the Islamic era (class 0) instances are misclassified as Abbasid (class 1), \textit{Aldoul wa al-emarat} (class 2), or Ottoman (class 3). By contrast, the Abbasid (class 1), \textit{Aldoul wa al-emarat} (class 2), and Ottoman (class 3) eras are identified most accurately. In general, misclassifications tend to occur between consecutive eras, reflecting the absence of sharp temporal boundaries.

\begin{figure}
\begin{center}
\includegraphics[width=0.45\textwidth]{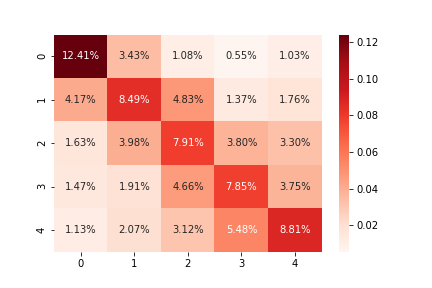}
\caption{The confusion matrix for the 300-years custom period classification using the OpenITI dataset and the ANN model.}
	\label{fig:cvopenitiann300}
\end{center}
\end{figure}

The confusion matrix for the 300-year custom periodization is presented in Figure \ref{fig:cvopenitiann300}. Class 0 (covering the pre-Islamic, Islamic, and first half of the Abbasid periods) is identified most accurately by the ANN. More than half of class 1 (the second half of the Abbasid and roughly 40\% of the \textit{Aldoul wa al-emarat} period) is misclassified, with most errors assigned to adjacent classes (the preceding or subsequent period).

\begin{table*}[]
	\centering
	\caption{The OpenITI dataset results with separating the authors using the RNNs model.}
	\begin{tabular}{@{}lcccc@{}}
		\toprule
		\textbf{Class Setup}                                                         & \multicolumn{1}{l}{\textbf{Accuracy}} & \multicolumn{1}{l}{\textbf{Precision}} & \multicolumn{1}{l}{\textbf{Recall}} & \multicolumn{1}{l}{\textbf{F1-Measure}} \\ \midrule
		\begin{tabular}[c]{@{}l@{}}Binary literature eras\end{tabular}        & 0.664                                 & 0.683                                  & 0.665                               & 0.656                                  \\ \hline
		3-eras                                                                  & 0.615                                 & 0.628                                  & 0.614                               & 0.615                                  \\ \hline
		\begin{tabular}[c]{@{}l@{}} Five literature eras \end{tabular}            & 0.400                                 & 0.415                                  & 0.403                               & 0.396                                  \\ \hline
		
		\begin{tabular}[c]{@{}l@{}}300-years custom period\\ (5-class)\end{tabular}  & 0.415                                 & 0.438                                  & 0.415                               & 0.382                                  \\ \hline
		\begin{tabular}[c]{@{}l@{}}200-years custom period\\ (8-class) \end{tabular} & 0.287                                 & 0.293                                  & 0.292                               & 0.282                                  \\ \hline
		\begin{tabular}[c]{@{}l@{}}100-years custom period\\ (15-class)\end{tabular} & 0.161                                 & 0.157                                  & 0.167                               & 0.157                                  \\ \bottomrule
	\end{tabular}
	\label{tab:openitirnnresults}
\end{table*}

\subsubsection{APCD Dataset Results}

Under author-disjoint splits, the APCD dataset attains its best results with the ANN model. Table \ref{tab:bestapcdresults} presents the ANN results, which outperform the alternative models. Lemmatization has a stronger positive impact on APCD than on OpenITI, likely owing to richer vocabulary and reduced sparsity.

Table \ref{tab:rnnapcdresults} reports the RNN results for APCD. The binary setup yields performance similar to OpenITI; however, for the remaining class setups, performance differs substantially, with the ANN outperforming the RNNs. Table \ref{tab:perclass-metrics-APCD} summarizes per-class metrics for the five-era classification. Overall performance is robust but uneven across classes (macro precision/recall/F1 $\approx$ 63.2/61.6/61.8\%). The Abbasid era is the best balanced (highest F1 $\approx$ 68.1\%, with precision 70.5\% and recall 65.8\%). The Islamic era shows the highest recall (75.8\%) but relatively low precision (56.3\%), suggesting liberal predictions and more false positives. The pre-Islamic era exhibits the opposite pattern—highest precision (68.2\%) but lowest recall (50.7\%)—indicating conservative predictions with many true samples missed. The Ottoman and Modern eras fall in the middle (F1 $\approx$ 60\% each), with the Modern era slightly weaker on recall (57.0\%).

\begin{table*}[]
	\centering
	\caption{The best results for the APCD dataset with separating the authors using the ANN model. The accuracy values are shown along with statistical significance interval at 95\% confidence level.}
	\begin{tabular}{@{}lcccc@{}}
		\toprule
		\textbf{Class Setup}                                                & \multicolumn{1}{l}{\textbf{Accuracy}} & \multicolumn{1}{l}{\textbf{Precision}} & \multicolumn{1}{l}{\textbf{Recall}} & \multicolumn{1}{l}{\textbf{F1-Measure}} \\ \midrule
		\begin{tabular}[c]{@{}l@{}}Binary literature eras \end{tabular}             & 0.783 $\pm{0.0143}$                                    & 0.784                                 & 0.782                              & 0.783                                  \\\hline
		\begin{tabular}[c]{@{}l@{}}Five literature eras \\ (with lemmatization)\end{tabular}              &0.654 $\pm{0.0225}$                                   & 0.661                                 & 0.647                               & 0.651                                  \\\hline
		
		11-Eras                                                                          & 0.358 $\pm{0.0177}$                                   & 0.346                                  & 0.352                               & 0.346                                  \\\hline
		12-Eras                                                                           & 0.191 $\pm{0.0167}$                                    & 0.189                                 & 0.184                               & 0.180                                \\ \bottomrule
	\end{tabular}
	\label{tab:bestapcdresults}
\end{table*}

\begin{table}[htbp]
\caption{Summary of the per-class performance for the 5-era classification using the APCD dataset.}
\label{tab:perclass-metrics-APCD}
\centering
\begin{tabular}{lcccc}
\hline
\textbf{Class} & \textbf{Accuracy (\%)} & \textbf{Precision (\%)} & \textbf{Recall (\%)} & \textbf{F1-score (\%)} \\
\hline
Pre-Islamic & 50.74 & 68.22 & 50.74 & 58.20 \\
Islamic & 75.76 & 56.33 & 75.76 & 64.62 \\
Abbasid & 65.81 & 70.50 & 65.81 & 68.07 \\
Ottoman & 58.61 & 61.25 & 58.61 & 59.90 \\
Modern & 57.00 & 59.74 & 57.00 & 58.34 \\
\hline
\end{tabular}
\end{table}

\begin{figure}
\begin{center}
\includegraphics[width=0.45\textwidth]{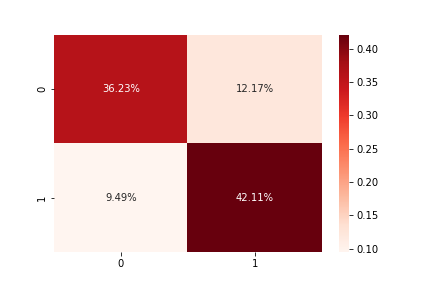}
\caption{The confusion matrix for the binary literature eras classification using the APCD dataset and the ANN model.}
	\label{fig:cv11c15lapcd}
\end{center}
\end{figure}

\begin{figure}
\begin{center}
\includegraphics[width=0.45\textwidth]{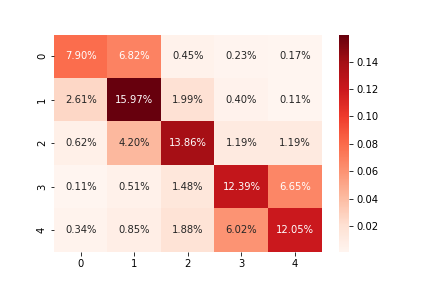}
\caption{The confusion matrix for the five literature eras classification using the APCD dataset and the ANN model.}
	\label{fig:cv5c15lapcd}
\end{center}
\end{figure}

Figure \ref{fig:cv11c15lapcd} shows the confusion matrix for the binary-era setup. One quarter of the Classical era (class 0) samples were predicted as Modern (class 1), whereas about 18\% of Modern (class 1) samples were predicted as Classical (class 0). Figure \ref{fig:cv5c15lapcd} presents the confusion matrix for the five-era setup. The Pre-Islamic era (class 0) exhibits the highest misclassification rate, while approximately 80\% of Islamic era (class 1) samples were correctly classified. The Ottoman and Modern eras are consecutive; about 30\% of Ottoman (class 3) samples were misclassified as Modern (class 4), and vice versa.

\begin{table*}[]
	\centering
	\caption{The APCD dataset results with separating the authors using the RNN model.}
	\begin{tabular}{@{}lcccc@{}}
		\toprule
		\textbf{Class Setup}                                                & \multicolumn{1}{l}{\textbf{Accuracy}} & \multicolumn{1}{l}{\textbf{Precision}} & \multicolumn{1}{l}{\textbf{Recall}} & \multicolumn{1}{l}{\textbf{F1-Measure}} \\ \midrule
		\begin{tabular}[c]{@{}l@{}}Binary literature eras \end{tabular}             & 0.759                                 & 0.759                                 & 0.757                           & 0.757                                  \\\hline
		\begin{tabular}[c]{@{}l@{}}Five literature eras\end{tabular}              &0.421                                & 0.464                                 & 0.404                               & 0.381                                  \\\hline
		
		11-Eras                                                                          & 0.213                                 & 0.235                                  & 0.234                               & 0.194                                  \\\hline
		12-Eras                                                                           & 0.135                                & 0.151                                 & 0.146                               & 0.132                                \\ \bottomrule
	\end{tabular}
	\label{tab:rnnapcdresults}
\end{table*}

\subsection{Temporal Classification with Authors Merged}
\subsubsection{OpenITI Dataset Results}

In addition to the author-disjoint results, results with merged authors are reported. As expected, era-classification performance generally improves when authors are merged—i.e., when the same authors appear in the validation and test sets as in the training set. This protocol helps isolate and illustrate the influence of authorial style on classification performance.

Across most class setups and in both datasets, the ANN outperforms the alternative models. The best OpenITI results are presented in Table \ref{tab:bestopenitiannresultsNS}, and the RNN results in Table \ref{tab:openitirnnresultsNS}. RNNs surpass the ANN only in the binary-era setup. The five-era configuration performs substantially better than the 300-year custom periodization, whereas the 200-year custom periodization yields results comparable to the 300-year setup.

\begin{table*}[]
	\centering
	\caption{The OpenITI dataset results for the ANN model with merging the authors.}
	\begin{tabular}{@{}lcccc@{}}
		\toprule         
  \textbf{Class Setup}     & \multicolumn{1}{l}{\textbf{Accuracy}} & \multicolumn{1}{l}{\textbf{Precision}} & \multicolumn{1}{l}{\textbf{Recall}} & \multicolumn{1}{l}{\textbf{F1-Measure}} \\ \midrule
		\begin{tabular}[c]{@{}l@{}}Binary literature eras \end{tabular}         & 0.829                                 & 0.829                                  & 0.829                              & 0.829                                  \\ \hline
		3-Eras                                                                   & 0.747                                 & 0.747                                  & 0.748                               & 0.747                                  \\ \hline
		\begin{tabular}[c]{@{}l@{}}Five literature eras  \end{tabular}            &  0.658                                 & 0.663                                  & 0.663                                & 0.663                                   \\ \hline
		
		\begin{tabular}[c]{@{}l@{}}300-years custom period\\ (5-class)\end{tabular}  & 0.548                                 & 0.549                                 & 0.550                              &  0.549                                   \\ \hline
		\begin{tabular}[c]{@{}l@{}}200-years custom period \\ (8-class)\end{tabular}  & 0.511                                & 0.516                                  & 0.516                                & 0.515                                  \\ \hline
		\begin{tabular}[c]{@{}l@{}}100-years custom period\\ (15-class)\end{tabular}  & 0.399                                &0.397                                  & 0.402                               & 0.399                                  \\ \bottomrule
	\end{tabular}
	\label{tab:bestopenitiannresultsNS}
\end{table*}

\begin{table*}[]
	\centering
	\caption{The OpenITI dataset results with merging the authors using the RNNs model.}
	\begin{tabular}{@{}lcccc@{}}
		\toprule
		\textbf{Class Setup}                                                           & \multicolumn{1}{l}{\textbf{Accuracy}} & \multicolumn{1}{l}{\textbf{Precision}} & \multicolumn{1}{l}{\textbf{Recall}} & \multicolumn{1}{l}{\textbf{F1-Measure}} \\ \midrule
		\begin{tabular}[c]{@{}l@{}}Binary literature eras \end{tabular}        & 0.830                                 & 0.830                                  & 0.830                              & 0.830                                  \\ \hline
		3-Eras                                                                   & 0.720                                 & 0.733                                  & 0.730                               & 0.725                                  \\ \hline
		\begin{tabular}[c]{@{}l@{}}Five literature eras  \end{tabular}            &  0.585                                 & 0.624                                  & 0.590                                & 0.596                                   \\ \hline
		
		\begin{tabular}[c]{@{}l@{}}300-years custom period\\ (5-class)\end{tabular}  & 0.458                                 & 0.476                                 & 0.459                              &  0.447                                   \\ \hline
		\begin{tabular}[c]{@{}l@{}}200-years custom period \\ (8-class)\end{tabular}  & 0.369                                & 0.412                                  & 0.374                                & 0.379                                  \\ \hline
		\begin{tabular}[c]{@{}l@{}}100-years custom period\\ (15-class)\end{tabular} & 0.204                                &0.212                                  & 0.210                               & 0.184                                  \\ \bottomrule
	\end{tabular}
	\label{tab:openitirnnresultsNS}
\end{table*}

\begin{figure}
\begin{center}
\includegraphics[width=0.45\textwidth]{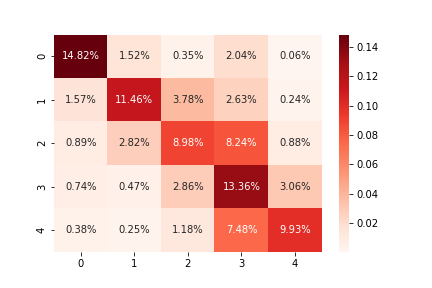}
\caption{Confusion matrix for five literature eras classification using the OpenITI dataset and the ANN model with merging the authors.}
	\label{fig:cvopenitins1}
\end{center}
\end{figure}

Figure \ref{fig:cvopenitins1} presents the confusion matrix for the five-era setup. The model correctly identified most samples from the Pre-Islamic (class 0) and Ottoman (class 1) eras. By contrast, roughly half of the \textit{Aldoul wa al-emarat} era (class 2) samples were misclassified as Ottoman (class 1).

\begin{figure}
\begin{center}
\includegraphics[width=0.45\textwidth]{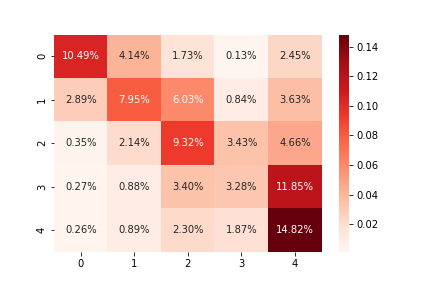}
\caption{Confusion matrix for 300-years custom period classification using the OpenITI dataset and the ANN model with merging the authors.}
	\label{fig:cvopenitins2}
\end{center}
\end{figure}

The confusion matrix for the 300-year custom periodization without separating authors is shown in Figure \ref{fig:cvopenitins2}. Class 3 is the most frequently misclassified period, whereas classes 0 and 4 are classified more accurately than the others. This pattern suggests a clearer distinction between classical and modern Arabic.

\subsubsection{APCD Dataset Results}
Table \ref{tab:bestapcdresultsNS} reports the APCD results for the ANN model under merged-author splits. ANN is the best-performing model for most APCD experiments. In the five-era setup, the strongest results are obtained with lemmatization.

RNN results for APCD are shown in Table \ref{tab:rnnapcdresults}. The binary setup improves when authors are merged, while most other class setups yield results similar to those obtained with author-disjoint splits.

Overall, OpenITI results improve with author merging, whereas APCD results are largely similar with and without author separation.

\begin{table*}[]
	\centering
	\caption{The best results for APCD with merging the authors using ANN model.}
	\begin{tabular}{@{}lcccc@{}}
		\toprule
		\textbf{Class Setup}                                            & \multicolumn{1}{l}{\textbf{Accuracy}} & \multicolumn{1}{l}{\textbf{Precision}} & \multicolumn{1}{l}{\textbf{Recall}} & \multicolumn{1}{l}{\textbf{F1-Measure}} \\ \midrule
		\begin{tabular}[c]{@{}l@{}}Binary literature\\ eras\end{tabular}           & 0.792                                &  0.792                                  & 0.791                               & 0.792                                  \\ \hline
		\begin{tabular}[c]{@{}l@{}}Five literature eras \\ (with lemmatization)\end{tabular}            & 0.698                               & 0.699                                  & 0.690                               & 0.691                                 \\ \hline
		
		11-eras                                                                        & 0.421                                & 0.417                                 & 0.403                             &0.401                               \\ \hline
		12-eras                                                                        &  0.258                                 & 0.235                                 & 0.241                              & 0.236                                \\ \bottomrule
	\end{tabular}
	\label{tab:bestapcdresultsNS}
\end{table*}

\begin{table*}[]
	\centering
	\caption{The APCD dataset results with merging the authors using RNN model.}
	\begin{tabular}{@{}lcccc@{}}
		\toprule
		\textbf{Class Setup}                                            & \multicolumn{1}{l}{\textbf{Accuracy}} & \multicolumn{1}{l}{\textbf{Precision}} & \multicolumn{1}{l}{\textbf{Recall}} & \multicolumn{1}{l}{\textbf{F1-Measure}} \\ \midrule
		\begin{tabular}[c]{@{}l@{}}Binary literature\\ eras\end{tabular}           & 0.785                                &  0.789                                  & 0.783                               & 0.783                                  \\ \hline
		\begin{tabular}[c]{@{}l@{}}Five literature eras \end{tabular}            & 0.448                               & 0.464                                  & 0.439                              & 0.390                                 \\ \hline
		
		11-eras                                                                        & 0.223                               & 0.150                                 & 0.232                             &0.162                               \\ \hline
		12-eras                                                                        &  0.170                                 & 0.198                                 & 0.165                              & 0.150                                \\ \bottomrule
	\end{tabular}
	\label{tab:rnnapcdresultsNS}
\end{table*}

\subsection{Result Analysis and Discussion}
This section investigates potential sources of error. The analysis focuses on the five-era setup for the OpenITI and APCD datasets, considering only the best results under the author-disjoint condition.

For OpenITI, the strongest results are obtained with the ANN model. In Figure \ref{fig:cvopenitiann5}, classes 0–4 correspond to Islamic, Abbasid, \textit{Aldoul wa al-emarat}, Ottoman, and Modern. Approximately half of the Islamic samples are misclassified as Abbasid or \textit{Aldoul wa al-emarat}. Similarly, for the Modern era, about half of the samples are misclassified as \textit{Aldoul wa al-emarat} or Ottoman. These patterns suggest overlap and similarity between consecutive eras. A plausible contributing factor is the dataset construction itself, as historical boundaries between eras are not sharply defined.

For APCD, the best performance is also achieved by the ANN model. In Figure \ref{fig:cv5c15lapcd}, classes 0–4 represent Pre-Islamic, Islamic, Abbasid, Ottoman, and Modern. The Islamic era is the most accurately predicted, whereas the Pre-Islamic era exhibits the highest misclassification rate—likely because it contains the fewest samples. In addition, more than a quarter of Ottoman samples are predicted as Modern and vice versa, indicating overlap and similarity between these neighboring eras where literary style may not change abruptly.

Word-frequency comparisons were conducted across eras. Although this analysis is purely statistical and does not account for context, it provides preliminary insight into the problem.

For selected samples, per-era word frequencies were computed to compare totals. Let a sample consist of a set of $n$ words, and let the classifier assign it to one era from a set of $m$ eras, as follows:
\begin{equation}
		S=\{w_0, w_1, ..., w_n\}\\
		E=\{e_0, e_1, ..., e_m\}\\
\end{equation}
where $S$ denotes the sample and $w_0$ denotes the first word in the sample; $E$ denotes the set of eras. For each era $e \in E$, the frequency of each word in the sample is counted as follows:

\begin{equation}
	 {Word frequency}=\sum_{i=0}^{n} w_{ie_j}
\end{equation}
where $n$ is the number of words in a sample, and $e_j$ is a specific era from the eras set. 

Table \ref{tab:samplesaapwf} illustrates four cases. In Case 1, the sample is misclassified and the (incorrect) predicted era also has the highest aggregate word frequency. In Case 2, the sample is correctly classified and the true era has the highest aggregate frequency. In Case 3, the sample is misclassified, and the highest aggregate frequency belongs to an era that is neither the true nor the predicted era. In Case 4, the sample is correctly classified, yet the highest aggregate frequency occurs in an era different from the true one.

\begin{table*}[]
	\centering
	\caption{The actual era, predicted era, and the words frequency per era for some samples from OpenITI datasets.}
	\begin{tabular}{@{}clllr@{}}
		\toprule
		\multicolumn{1}{l}{\textbf{Sample ID}} & \textbf{Actual Era}                                                                          & \textbf{Predicted Era}                                                                         & \textbf{Era}                                                                          & \multicolumn{1}{l}{\textbf{\begin{tabular}[c]{@{}l@{}}Word Frequency\end{tabular}}} \\ \midrule
		\multirow{5}{*}{1}                     & \multirow{5}{*}{Modern}                                                                      & \multirow{5}{*}{\begin{tabular}[c]{@{}l@{}}\textit{Aldoul wa} \\ \textit{ al-emarat}\end{tabular}} & Islamic                                                                               & 58,455                                                                                         \\ \cline{4-5}
		&                                                                                              &                                                                                                & Abbasid                                                                               & 113,925                                                                                          \\ \cline{4-5}
		&                                                                                              &                                                                                                & \textbf{\begin{tabular}[c]{@{}l@{}}\textit{Aldoul wa} \\ \textit{ al-emarat}\end{tabular}} & \textbf{117,467}                                                                               \\\cline{4-5}
		&                                                                                              &                                                                                                & Ottoman                                                                               & 75,741                                                                                         \\ \cline{4-5}
		&                                                                                              &                                                                                                & Modern                                                                                & 93,003                                                                                           \\ \hline
		\multirow{5}{*}{2}                     & \multirow{5}{*}{Abbasid}                                                                     & \multirow{5}{*}{Abbasid}                                                                       & Islamic                                                                               & 99,317                                                                                        \\ \cline{4-5}
		&                                                                                              &                                                                                                & \textbf{Abbasid}                                                                      & \textbf{150,765}                                                                               \\ \cline{4-5}
		&                                                                                              &                                                                                                & \begin{tabular}[c]{@{}l@{}}\textit{Aldoul wa} \\ \textit{ al-emarat}\end{tabular}           & 147,992                                                                                          \\ \cline{4-5}
		&                                                                                              &                                                                                                & Ottoman                                                                               & \multicolumn{1}{r}{115,974}                                                                      \\ \cline{4-5}
		&                                                                                              &                                                                                                & Modern                                                                                & \multicolumn{1}{r}{119,782}                                         \\ \hline
		\multirow{5}{*}{3}                     & \multirow{5}{*}{Ottoman}                                                                     & \multirow{5}{*}{\begin{tabular}[c]{@{}l@{}}\textit{Aldoul wa} \\ \textit{ al-emarat}\end{tabular}}  & Islamic                                                                               &  71,406                                                                                        \\ \cline{4-5}
		&                                                                                              &                                                                                                & \textbf{Abbasid}                                                                      & \textbf{132,326}                                                                               \\ \cline{4-5}
		&                                                                                              &                                                                                                & \begin{tabular}[c]{@{}l@{}}\textit{Aldoul wa} \\ \textit{ al-emarat}\end{tabular}           & 126,299                                                                                        \\ \cline{4-5}
		&                                                                                              &                                                                                                & Ottoman                                                                               & 85,586                                                                                         \\ \cline{4-5}
		&                                                                                              &                                                                                                & Modern                                                                                & 95,307                                                                                         \\ \hline
		\multirow{5}{*}{4}                     & \multirow{5}{*}{\begin{tabular}[c]{@{}l@{}}\textit{Aldoul wa} \\ \textit{ al-emarat}\end{tabular}} & \multirow{5}{*}{\begin{tabular}[c]{@{}l@{}}\textit{Aldoul wa} \\ \textit{ al-emarat}\end{tabular}}  & Islamic                                                                               & 54,052                                                                                         \\ \cline{4-5}
		&                                                                                              &                                                                                                & \textbf{Abbasid}                                                                      & \textbf{65,135}                                                                                \\ \cline{4-5}
		&                                                                                              &                                                                                                & \begin{tabular}[c]{@{}l@{}}\textit{Aldoul wa} \\ \textit{ al-emarat}\end{tabular}           & 65,056                                                                                          \\ \cline{4-5}
		&                                                                                              &                                                                                                & Ottoman                                                                               & 56,480                                                                                         \\ \cline{4-5}
		&                                                                                              &                                                                                                & Modern                                                                                &  52,854                                                                                                                    \\ \bottomrule
	\end{tabular}
	\label{tab:samplesaapwf}
\end{table*}

Word occurrences were counted per era as described for the OpenITI analysis. Analogously, four cases for APCD are presented in Table \ref{tab:samplesaapwfapcd}. In the first sample, the incorrectly predicted era has the highest aggregate word frequency. In the second, the sample is correctly classified and the true era has the highest total. The remaining two cases show the converse patterns: in the third, the sample is misclassified but the highest total belongs to the true era; in the fourth, the sample is misclassified and the highest total belongs to an era different from both the true and the predicted labels.

From Table \ref{tab:samplesaapwfapcd}, it is evident that raw word counts are not always decisive for a given era. For example, in the final case—where the true era is Pre-Islamic—the number of words associated with that era is lower than in others, indicating that contextual usage outweighs simple frequency.

Across both datasets, total word occurrences can influence predictions, but not deterministically. Several cases show no direct correspondence between the highest aggregate frequency and the predicted label. Consequently, classification depends on more than total word counts alone.

\begin{table*}[]
	\centering
	\caption{The actual era, predicted era, and the words frequency per era for some samples from APCD dataset}
	\begin{tabular}{@{}clllr@{}}
		\toprule
		\multicolumn{1}{l}{\textbf{Sample ID}} & \textbf{Actual Era}          & \textbf{Predicted Era}   & \textbf{Era}     & \multicolumn{1}{l}{\textbf{\begin{tabular}[c]{@{}l@{}}Word Frequency\end{tabular}}} \\ \midrule
		\multirow{5}{*}{1}                     & \multirow{5}{*}{Modern}      & \multirow{5}{*}{Ottoman} & Pre-Islamic      & 24,296                                                                                         \\ \cline{4-5}
		&                              &                          & Islamic          & 45,832                                                                                         \\ \cline{4-5}
		&                              &                          & Abbasid          & 51,443                                                                                         \\ \cline{4-5}
		&                              &                          & \textbf{Ottoman} & \textbf{53,749}                                                                                \\ \cline{4-5}
		&                              &                          & Modern           & 49,004                                                                                           \\ \hline
		\multirow{5}{*}{2}                     & \multirow{5}{*}{Abbasid}     & \multirow{5}{*}{Abbasid} & Pre-Islamic      & 21,783                                                                                         \\
		&                              &                          & Islamic          & 40,559                                                                                         \\ \cline{4-5}
		&                              &                          & \textbf{Abbasid} & \textbf{43,822}                                                                                \\ \cline{4-5}
		&                              &                          & Ottoman          & 43,527                                                                                         \\ \cline{4-5}
		&                              &                          & Modern           & 40,465                                                                                         \\  \hline
		\multirow{5}{*}{3}                     & \multirow{5}{*}{Ottoman}     & \multirow{5}{*}{Modern} & Pre-Islamic      & 25,253                                                                                         \\ \cline{4-5}
		&                              &                          & Islamic          & 46,818                                                                                           \\ \cline{4-5}
		&                              &                          & Abbasid & 51,990                                                                                \\ \cline{4-5}
		&                              &                          & \textbf{Ottoman}        & \textbf{53,766}K                                                                                         \\ \cline{4-5}
		&                              &                          & Modern           & 49,434                                                                                         \\  \hline
		\multirow{5}{*}{4}                     & \multirow{5}{*}{Pre-Islamic} & \multirow{5}{*}{Islamic} & Pre-Islamic      & 18,847                                                                                         \\ \cline{4-5}
		&                              &                          & Islamic          & 34,724                                                                                         \\ \cline{4-5}
		&                              &                          & \textbf{Abbasid} & \textbf{35,922}                                                                                \\ \cline{4-5}
		&                              &                          & Ottoman          & 34,143                                                                                         \\ \cline{4-5}
		&                              &                          & Modern           & 31,457                                                                                         \\ \bottomrule
	\end{tabular}
	\label{tab:samplesaapwfapcd}
\end{table*}

An additional point is that literary change is gradual and boundaries are inherently porous, as reflected in the confusion matrices (Figures 3 and 6). To quantify this, performance was recomputed after merging adjacent eras. For OpenITI, accuracy increases from 43.6\% to 76.6\%; for the poetry dataset, from 65.4\% to 93.1\%. These gains support the view that literary evolution forms a continuum rather than a sequence of sharply separated periods.

\subsection{Comparison to Other Works in The Literature}
\label{comare}

As no other Arabic-domain studies were identified besides \citet{orabi2020classical}, their binary classification task was replicated to enable an objective comparison. The comparison uses their dataset. Because the RNN model outperformed the ANN on the poetry dataset’s binary task, the word-level RNN was selected for this comparison. Table \ref{tab:comparcnnvsRNNs} reports the results of the RNN alongside the CNN of \citet{orabi2020classical}. The data setup used here is more restricted than theirs, as described in Section \ref{sec:datasets}. \citet{orabi2020classical} split the data into training and validation sets and evaluated directly on the validation set. The word-level RNN achieves higher validation performance; statistical significance tests on accuracy indicate that these validation results are superior. However, the test-set results are comparable to their validation results, with no significant difference.

\begin{table*}[]
	\centering
	\caption{Comparison between our best model results and the CNN model proposed by Orabi et al. \cite{orabi2020classical}.}
	\begin{tabular}{@{}llccc@{}}
		\toprule
		\textbf{Model}           & \textbf{\begin{tabular}[c]{@{}l@{}}Experiment\\ Configuration\end{tabular}}                                    & \textbf{Accuracy} & \textbf{F1-Measure} & \textbf{\begin{tabular}[c]{@{}l@{}}Statistical\\ Significance\end{tabular}} \\ \midrule
		\multirow{2}{*}{Our RNN} & \multirow{2}{*}{\begin{tabular}[c]{@{}l@{}}Training (73\%) -\\ Validation (12\%) - \\Test (15\%)\end{tabular}} & \makecell[c]{0.92 \\ Validation}              & \makecell[c]{0.917 \\Validation}              & ± 0.0045                                                                    \\ \cmidrule(l){3-5} 
		&                                                                                                                & \makecell[c]{0.91 \\ Test}             & \makecell[c]{0.911 \\ Test}             & ± 0.0042                                                                    \\ \midrule
		\makecell[l]{CNN by\\
		Orabi et al. \cite{orabi2020classical}}                   & \begin{tabular}[c]{@{}l@{}}Training (90\%) - \\ Test (10\%)\end{tabular}                                 & 0.91              & 0.910              & ± 0.0052                                                                    \\ \bottomrule
	\end{tabular}
	\label{tab:comparcnnvsRNNs}
\end{table*}

\subsection{Summary and Implications of the Work}
\label{ssec:implications}

The main findings of this work can be summarized as follows:
\begin{enumerate}
\item Automatic era classification of Arabic text, although difficult, is feasible. Accuracy depends on several factors, including input length: the more text and context available, the more accurate the prediction.
\item It appears easier to identify the Abbasid, \textit{Aldoul wa al-emarat}, and Ottoman eras than the Islamic and Modern eras. Moreover, most eras overlap primarily with neighboring eras because their temporal boundaries are not sharply defined.
\item Era identification from \emph{poetry} texts is relatively easier than from \emph{non-poetry} literature. Lemmatization is particularly beneficial for poetry, likely due to its richer vocabulary and greater sparsity; lemmatization helps keep the vocabulary size manageable.
\item For non-poetry literature, including the same authors in both the training and test sets improves results, indicating that authorial style influences era classification. In poetry, however, merging authors does not materially affect the results.
\end{enumerate}

The main implications of the work are as follows:
\begin{enumerate}
\item Automated era classification enables analysis of trends, themes, and stylistic elements characteristic of specific periods, aiding understanding of the evolution of Arabic literature.
\item Because literary texts reflect social, cultural, and political contexts, automated classification can assist in examining shifts in norms, ideologies, and historical events across eras.
\item Automation facilitates large-scale analysis of literary corpora, supporting research on authorship attribution, genre, and linguistic evolution over time.
\item Identifying and categorizing Arabic literature by era contributes to preserving cultural heritage by ensuring that works from different periods are systematically documented and archived.
\end{enumerate}

\section{Conclusions}
\label{sec:conc}

Languages evolve over time and are affected by diverse changes, such as semantic and lexical shifts. Numerous linguistic studies document the historical development of Arabic. Historians and linguists commonly divide Arabic literature into eras corresponding to distinct time periods. Automatic labeling of texts by period is useful in linguistics and history \cite{niculae2014temporal}. This paper leverages modern machine-learning techniques to classify Arabic texts into various eras and time periods.

Two publicly available corpora were used: OpenITI for literary prose and APCD for poetry. Both span from the Pre-Islamic to the modern period. Multiple class setups were examined, including binary and five-era configurations for each dataset. The primary models were ANNs and RNNs. ANNs achieved better results on both OpenITI and APCD in most setups, whereas for binary classification the RNN outperformed some baselines, including the model of Orabi et al. \cite{orabi2020classical}. For example, on the five-class OpenITI task, accuracy ranges from 0.40 to 0.66—above chance yet far from perfect—underscoring the task’s difficulty and the need for further research.

In addition to the class setups, several experimental settings were evaluated. Removing stop words did not improve performance. Lemmatization improved results only for APCD (poetry), suggesting that poems’ richer and more varied vocabulary benefits from vocabulary normalization. Performance improved in most experiments after merging authors across splits.

The proposed machine-learning approaches consistently outperform random guessing. Orabi et al. \cite{orabi2020classical} reported their best results with a CNN, while the RNN matched or exceeded those results depending on the evaluation split. Under merged-author splits on OpenITI, historically defined era setups performed better than custom time bins; the five-era scheme appears to capture literary change more faithfully. Custom bins revealed gaps between some adjacent periods (treating them as discrete steps), whereas era-based schemes reflected overlapping transitions.

Future work includes improving lemmatization for Classical Arabic, which proved less effective in the present experiments; such advances could enhance era classification. Another direction is to evaluate pretrained transformers (e.g., AraBERT, MARBERT) for temporal classification, while accounting for the fact that many Arabic transformers are trained predominantly on modern-era data and may introduce temporal bias.

Incorporating external knowledge about named persons mentioned in the text may also help, for example by applying named-entity recognition to extract person names and linking them to biographical resources.

Colgrove et al. \cite{colgrove2010literary} focused on influential authors associated with each period in their dataset; a similar approach could be explored for Arabic literature, though identifying period-leading authors remains a debated topic among historians and literary critics.

Belinkov et al. \cite{belinkov2019studying} used an efficient text-reuse detection algorithm and reported that approximately 20\% of OpenITI consists of reused material (e.g., extremely frequent short phrases such as \textit{peace be upon him} “\<عليه السلام>”). They released a modified version of OpenITI with reused passages removed. Future work could examine how using this de-reused corpus affects temporal classification performance.

\section*{Acknowledgements}
The authors would like to thank King Fahd University of Petroleum and Minerals (KFUPM) for supporting this work. Irfan Ahmad would like to additionally thank Saudi Data and AI Authority (SDAIA) and KFUPM for supporting him through SDAIA-KFUPM Joint Research Center for Artificial Intelligence grant number JRC--AI--RFP--10.

\bibliographystyle{apacite}
\bibliography{refs}

\newpage
\appendix

\section{Pilot Experiments Using Some Other Models}
\begin{figure}
     \centering
     \includegraphics[width=1.5in]{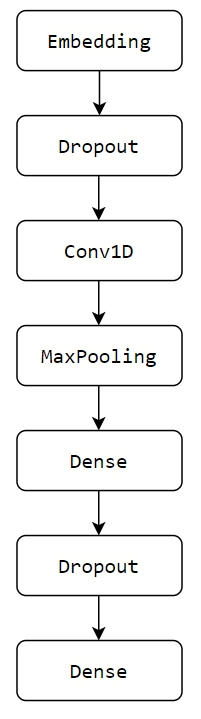}
     \caption{The architecture of the CNN model applied by Orabi et al. \cite{orabi2020classical}}
     \label{fig:CNN22s}
    \end{figure}

In addition to the ANN and RNN models, pilot experiments were conducted with the following:
\begin{itemize}
   
    \item \textbf{CNN:} Is mostly used with images and objects classification, recognition, and detection. It achieved exceptional classification performance in the literature \cite{ozyurt2020expert}. Recently CNN has acquired an interest in text classification problems. Orabi et al. \cite{orabi2020classical} applied it to classify Arabic poems into eras \cite{orabi2020classical}. Therefore, we will follow the same architecture they used in their experiments. Figure \ref{fig:CNN22s} shows the architecture for the CNN model that been applied in \cite{orabi2020classical}. They used words as tokens as input for the model that feed into the embedding layer. Moreover, they used a convolution layer of one dimension (Conv1D). Also, they applied a max-pooling layer of one dimension.

\item \textbf{Logistic Regression Model}
Logistic regression is one of the traditional machine learning algorithms that is used mostly with classification problems. Logistic regression uses \textit{sigmoid} function to calculate the probability of the distance between the input and decision boundary. Thus, the output will a value between zero and one.

The cost function for logistic regression is Log Loss. Log Loss is the classification metric based on probabilities, a lower cost value indicates better predictions. The formula for the Log loss is defined as follows:
\begin{equation}
    \text{Log Loss}=\sum_{(x,y) \in D} -y\log ( \hat{y})-(1-y)-(1- \hat{y})
\end{equation}

where $(x,y) \in D$ is the dataset which contains $x$ as sample and $y$ as label, $y$ is the label, and $\hat{y}$ is the predicted value.

Furthermore, regularization is important for logistic regression model to avoid overfitting issue. Ridge regularization is added to the objective. The ridge penalty is equal to $\lambda \lVert w \rVert_2 ^2$. The hyperparameter $\lambda$ is used to the control the penalty on the weights. Larger $\lambda$ imposes stronger shrinkage and generally reduces overfitting. The hyperparameter $\lambda$ is tuned, and the selected value is reported with the results. Optimization employs the limited-memory Broyden–Fletcher–Goldfarb–Shanno (\textit{LBFGS}) algorithm.
    
\end{itemize}

\subsection{Experiments and Results}
In addition to the ANN and RNN models, pilot experiments were conducted with the standard five-era setup on both the OpenITI and APCD datasets. For APCD, the configuration with 15 verses per sample was used, as it yielded the best results.

Logistic regression was implemented with {\itshape scikit-learn}. The hyperparameter \textit{C} represents the inverse of $\lambda$; accordingly, \textit{C} was tuned for each dataset and the best value reported.

For the CNN, the architecture of \citet{orabi2020classical} was followed. However, the data split differs from theirs: the present experiments use the three-way split described in Section \ref{sec:datasets}. Both Adam and RMSProp optimizers were evaluated, and the better performer is reported for each dataset. The CNN batch size was 128.

\subsubsection{OpenITI Dataset}
For logistic regression, \textit{C} was tuned to reduce overfitting; the best performance was obtained with \textit{C} = 0.001. For the CNN, two optimizers were examined; RMSProp outperformed Adam.

Results are summarized in Table \ref{tab:othermodelsopeniti}. Performance was similar and not affected by stop-word removal; for example, CNN accuracy remained 0.46 both before and after removing stop words.

\begin{figure}[ht!]
\centering
\subfloat[Logistic regression model \label{sub:cvopenitiother1}]{\includegraphics[height=2in]{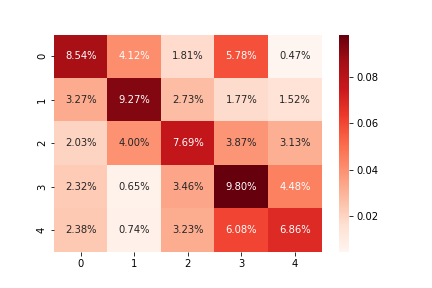}}
\subfloat[CNN model \label{sub:cvopenitiother2}]{\includegraphics[height=2in]{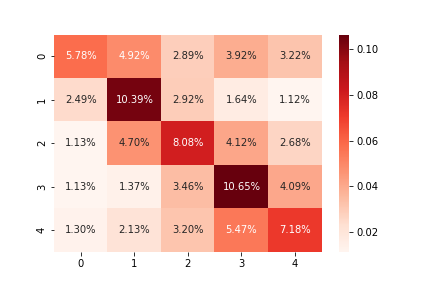}}
\caption{Confusion matrix for five literature eras setup using logistic regression (A) and CNN (B) models for the OpenITI dataset.}
  \label{fig:cvopenitiother}
\end{figure}

Figure \ref{fig:cvopenitiother} presents confusion matrices for the five-era setup using logistic regression and CNN. In Figure \ref{fig:cvopenitiother}(a) (logistic regression), approximately half of the Islamic era (class 0) samples are misclassified; notably, about one quarter are assigned to the Ottoman era (class 3), which is not adjacent to Islamic. In Figure \ref{fig:cvopenitiother}(b) (CNN), roughly half of the Abbasid (class 1) and Ottoman (class 3) samples are correctly classified, while only about one quarter of Islamic era (class 0) samples are identified correctly.

\begin{table}[htbp]
\caption{Summary of the results for the OpenITI dataset using CNN and Logistic Regression models.}
\label{tab:othermodelsopeniti}
\centering
\begin{tabular}{@{}l l cc@{}}
\toprule
Model & Remove Stop-words & Validation Accuracy & Test Accuracy \\
\midrule
\multirow{2}{*}{CNN}                 & No  & 0.46 & 0.46 \\
                                     & Yes & 0.46 & 0.42 \\
\addlinespace
\multirow{2}{*}{Logistic Regression} & No  & 0.46 & 0.42 \\
                                     & Yes & 0.45 & 0.41 \\
\bottomrule
\end{tabular}
\end{table}

\subsubsection{APCD}
The hyperparameter \textit{C} for logistic regression was tuned based on validation results; the best performance was obtained with \textit{C} = 0.01. For the CNN model, multiple optimizers were evaluated; unlike OpenITI, the Adam optimizer outperformed RMSProp.

Table \ref{tab:othermodelsapcd} presents results for the additional models on the APCD dataset. For logistic regression, performance was similar before and after stop-word removal. For the CNN model, results were better without removing stop words; for example, accuracy was 0.51 before removal and 0.49 after.

\begin{figure}[ht!]
\centering
\subfloat[Logistic regression model \label{sub:cvapcdother1}]{\includegraphics[height=2in]{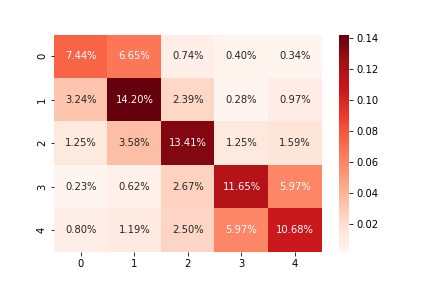}}
\subfloat[CNN model \label{sub:cvapcdother2}]{\includegraphics[height=2in]{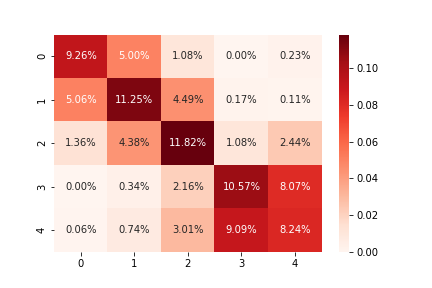}}
\caption{Confusion matrix for five literature eras setup using logistic regression (A) and CNN (B) models for the APCD dataset.}
  \label{fig:cvapcdother}
\end{figure}

Figure \ref{fig:cvapcdother}(a) shows the confusion matrix for the five-era setup using logistic regression. The model correctly predicted most Islamic (class 1) samples, whereas about half of the Pre-Islamic (class 0) samples were misclassified. Figure \ref{fig:cvapcdother}(b) presents the confusion matrix for the five-era setup using the CNN model. Approximately half of Ottoman (class 3) samples were classified as Modern (class 4), and \emph{vice versa}. In addition, 15\% of Modern (class 4) samples were predicted as Abbasid (class 2).

Although the Pre-Islamic era (class 0) contains relatively few samples, logistic regression performed better than CNN on this class. Conversely, CNN classified the Abbasid era (class 2) more accurately than logistic regression. CNN also tended to confuse the final two eras with each other.

\begin{table}[htbp]
\caption{Summary of the results for the APCD dataset using CNN and Logistic Regression models.}
\label{tab:othermodelsapcd}
\centering
\begin{tabular}{@{}l l cc@{}}
\toprule
Model & Remove Stop-words & Validation Accuracy & Test Accuracy \\
\midrule
\multirow{2}{*}{CNN}                 & No  & 0.50 & 0.51 \\
                                     & Yes & 0.48 & 0.46 \\
\addlinespace
\multirow{2}{*}{Logistic Regression} & No  & 0.60 & 0.57 \\
                                     & Yes & 0.61 & 0.57 \\
\bottomrule
\end{tabular}
\end{table}

\end{document}